\documentclass[10pt,twocolumn,letterpaper]{article}
\PassOptionsToPackage{table}{xcolor}
\usepackage{cvpr}

\usepackage[accsupp]{axessibility}  
\usepackage{algorithm}
\usepackage{algorithm,algpseudocode}
\usepackage{algorithmicx}
\usepackage{graphicx}
\usepackage{amsmath}
\usepackage{amssymb}
\usepackage{booktabs}
\usepackage[utf8]{inputenc} 
\usepackage[T1]{fontenc}  
\usepackage{url}           
\usepackage{booktabs}       
\usepackage{amsfonts}     
\usepackage{nicefrac}      
\usepackage{microtype}      
\usepackage{mathtools}
\usepackage{arydshln}
\usepackage{amsthm}
\usepackage{bm}
\usepackage{enumitem}
\usepackage{multirow}
\usepackage{cuted}

\usepackage{multirow}
\usepackage{arydshln}

\definecolor{tabfirst}{rgb}{1, 0.7, 0.7} 
\definecolor{tabsecond}{rgb}{1, 0.85, 0.7} 
\definecolor{tabthird}{rgb}{1, 1, 0.7}
\definecolor{cvprblue}{rgb}{0.21,0.49,0.74}
\usepackage[pagebackref,breaklinks,colorlinks,allcolors=cvprblue]{hyperref}

\newcommand{\bx}{\mathbf{x}}
\newcommand{\by}{\mathbf{y}}

\newcommand{\indep}{\mathrel{\perp\!\!\!\perp}}

\def\paperID{****}
\def\confName{CVPR}
\def\confYear{2025}

\title{SyncSDE: A Probabilistic Framework for Diffusion Synchronization}

\author{
Hyunjun Lee\textsuperscript{1}\thanks{Equal Contributions.} \hspace{10mm} Hyunsoo Lee\textsuperscript{1}\footnotemark[1]  \hspace{10mm} Sookwan Han\textsuperscript{1,2}\footnote[2] \\ \\ \\
   \textsuperscript{1}ECE, Seoul National University \hspace{10mm} \textsuperscript{2}Republic of Korea Air Force \\
   {\tt\small \{hjl1013,\,philip21,\,jellyheadandrew\}@snu.ac.kr}\vspace{-4em}
}

\begin{document}
\maketitle
\footnotetext[2]{Project Lead.}
\vspace{-10pt}
\begin{strip}
\centering
\includegraphics[width=\linewidth]{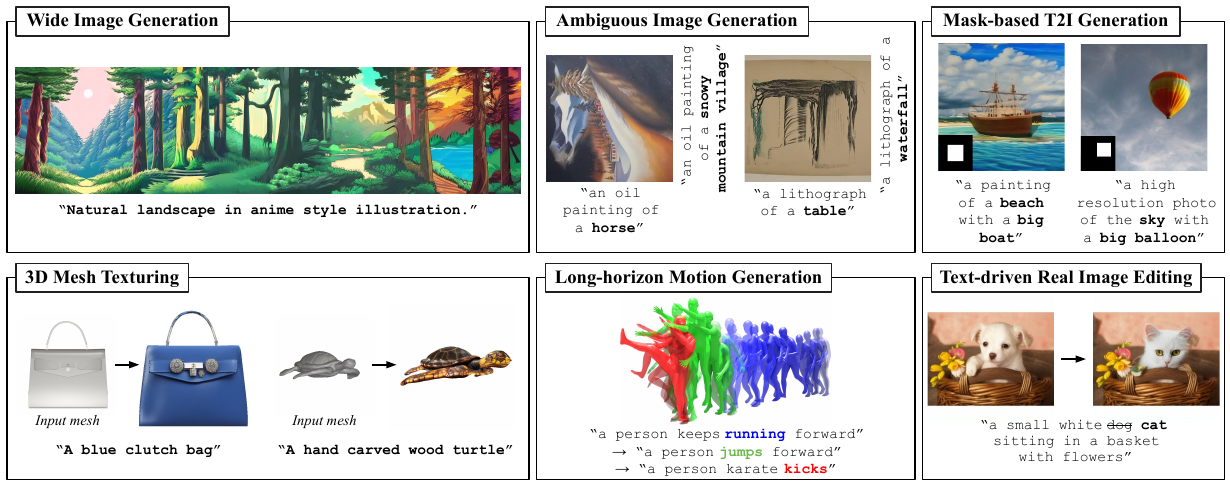} 
\captionof{figure}{
\textbf{Applications of SyncSDE.} SyncSDE analyzes diffusion synchronization to identify where the correlation strategies should be focused, enabling coherent and high-quality results across diverse collaborative generation tasks.
\label{fig:teaser}
\vspace{-10pt}
}
\end{strip}

\begin{abstract}
There have been many attempts to leverage multiple diffusion models for collaborative generation, extending beyond the original domain. A prominent approach involves synchronizing multiple diffusion trajectories by mixing the estimated scores to artificially correlate the generation processes. However, existing methods rely on naive heuristics, such as averaging, without considering task specificity. These approaches do not clarify why such methods work and often produce suboptimal results when a heuristic suitable for one task is blindly applied to others. In this paper, we present a probabilistic framework for analyzing why diffusion synchronization works and reveal where heuristics should be focused—modeling correlations between multiple trajectories and adapting them to each specific task. We further identify optimal correlation models per task, achieving better results than previous approaches that apply a single heuristic across all tasks without justification.
\end{abstract}   

\section{Introduction}
\label{sec:introduction}

Diffusion models~\cite{ho2020denoising, song2020denoising, song2023score, ho2022cascaded} have achieved remarkable success in generating high-quality images~\cite{rombach2022high, saharia2022photorealistic, ramesh2022hierarchical}, 3D scenes~\cite{liu2024pyramid, tang2024diffuscene, zhai2023commonscenes}, human and motion~\cite{tevet2023human, yuan2023physdiff, tevet2024closd, han2023chorus, kim2024beyond, kim2023chupa}, and videos~\cite{levon2023text2video, yaniv2023sinfusion, ho2022video}.
Despite their success, these models are typically trained on fixed-domain data, limiting their ability to generate diverse data formats (e.g., varying shapes or dimensions).
This constraint reduces the flexibility of diffusion models and narrows the range of generative tasks they can effectively handle.

To harness the diverse generative capabilities of multiple diffusion models with varying characteristics, existing approaches~\cite{jaihoon2024synctweedies, daniel2024visual, omer2023multidiffusion, geng2024factorized} use heuristics to synchronize diffusion trajectories, ensuring consistency across the generations managed by each trajectory.
For instance, Visual Anagrams~\cite{daniel2024visual} generates images with optical illusions from different perspectives, while SyncTweedies~\cite{jaihoon2024synctweedies} explores multiple heuristics to align generation paths, enabling the creation of panoramic images and even 3D textures.

Although previous approaches with diverse synchronization heuristics demonstrate promising results in collaborative generation tasks, they do not explain why synchronization works, relying solely on empirical evidence.
This lack of theoretical grounding limits both inter-task and intra-task generalizability, leading to inconsistent performance across tasks.
As a result, users must experiment extensively to find optimal synchronization strategies for each new task, hindering the scalability of using multiple diffusion models beyond familiar scenarios.
For instance, SyncTweedies~\cite{jaihoon2024synctweedies} tests 60 different synchronization strategies to approximate optimal results for given tasks.
Repeating this process for each new compositional generation task would severely limit the practical use of multiple diffusion models—especially without theoretical support to validate whether the results are truly optimal.

Our paper addresses the \textit{why} behind synchronization by introducing a probabilistic framework that formulates it as the optimization of two distinct terms.
In particular, one term models the correlation between diffusion trajectories, providing a foundation for applying human heuristics as strategic choices.
Supported by theoretical analysis, we investigate which synchronization strategies yield the best results across both existing and novel tasks, showing that naive application of heuristics often leads to suboptimal outcomes.
This work is the first to analyze \textit{why} synchronization works and to leverage this understanding to guide \textit{where} strategy selection should be focused for future tasks.
We demonstrate scalability by applying our method to a wide range of tasks and show that, while naive strategies frequently fall short, our approach consistently achieves superior results.
We refer to this method as \textbf{SyncSDE}: Synchronization of Stochastic Differential Equations.
The main contributions of our work are summarized as follows:

\begin{itemize}[label=$\bullet$]
    \item We introduce a probabilistic framework for diffusion synchronization, providing a theoretical foundation to understand \textit{why} synchronization works.
    \item Our approach reduces redundant empirical testing by identifying \textit{where} heuristics should be applied, mitigating the suboptimal outcomes of existing naive strategies.
    \item Extensive experiments across diverse diffusion synchronization tasks demonstrate the superior performance of our method over state-of-the-art baselines, reinforcing its generalizability to novel tasks.
\end{itemize}


\section{Related work}
\label{sec:related_work}

Recent advances in diffusion models have unlocked a wide range of applications, powered by foundation models such as Stable Diffusion~\cite{rombach2022high}, DeepFloyd~\cite{DeepFloydIF}, ControlNet~\cite{zhang2023adding}.
Numerous studies build on these pretrained models to tackle specific tasks, including compositional generation.

\noindent \textbf{Diffusion Models.} Diffusion models generate realistic images by progressively denoising Gaussian noise. DDPM~\cite{ho2020denoising} and DDIM~\cite{song2020denoising} implement this process via discrete sampling, which is known to approximate stochastic differential equations (SDEs)~\cite{song2023score}, forming the theoretical basis for diffusion models.
Latent diffusion models~\cite{rombach2022high} improve efficiency by operating in latent space, with Stable Diffusion being the most widely used. Pixel-based models like DeepFloyd~\cite{DeepFloydIF} are also gaining attention.
Beyond image synthesis, diffusion models are applied to tasks such as image-to-image translation~\cite{bahjat2023imagic, su2022dual, meng2021sdedit, dong2023prompt, cao2023masactrl, mokady2023null}, human motion generation~\cite{tevet2023human,tevet2024closd,yuan2023physdiff} where models edit target details while preserving source structure.
For instance, Imagic~\cite{bahjat2023imagic} fine-tunes pretrained models for this purpose.

\noindent \textbf{Diffusion Synchronization.}
Diffusion synchronization enables collaborative generation by synchronously sampling from multiple diffusion trajectories while maintaining consistency across them. It extends the capabilities of a single diffusion model to support tasks such as generating images of arbitrary sizes~\cite{omer2023multidiffusion, jaihoon2024synctweedies, zhang2023diffcollage, wang2023360}, creating seamless textures~\cite{liu2024text, richardson2023texture, bensadoun2024meta, chen2023text2tex, youwang2024paintit, jaihoon2024synctweedies}, producing optical illusions~\cite{daniel2024visual, jaihoon2024synctweedies, geng2024factorized}, and synthesizing complex motions~\cite{zhang2023diffcollage}.
By leveraging the prior knowledge encoded in pretrained diffusion models, these methods require no additional training, expanding applicability across diverse domains.
For example, SyncTweedies~\cite{jaihoon2024synctweedies} addresses a range of synchronization tasks by empirically testing 60 strategies, ultimately adopting an averaging method based on Tweedie’s formula~\cite{stein1981estimation}.
Other works target specific tasks: MultiDiffusion~\cite{omer2023multidiffusion} focuses on wide image and mask-based T2I generation using bootstrapping for improved localization; Visual Anagram~\cite{daniel2024visual} produces ambiguous images that shift with view transformations; and SyncMVD~\cite{liu2024text} generates UV texture maps from 3D meshes and text prompts.
These methods synchronize trajectories by averaging intermediate signals—such as predicted noise or latents—but offer no theoretical justification for why this works.
In contrast, we propose a probabilistic framework that explicitly models correlations between diffusion trajectories, providing the first theoretical foundation for diffusion synchronization.

\section{Method}
\label{sec:method}

\subsection{Preliminaries}
\subsubsection{Diffusion sampling}

Starting from Gaussian noise $p_T \sim \mathcal{N}(\mathbf{0}, \mathbf{I})$, the DDIM~\cite{song2020denoising} reverse process samples $\mathbf{x}_T \sim p_T$ and then sequentially samples $\mathbf{x}_{t-1}$ from $\mathbf{x}_t$ using the distribution:
\begin{align}
q_{\sigma}&(\bx_{t-1}|\bx_{t}, \bx_0)  \\ 
= & \; \mathcal{N} (\sqrt{\alpha_{t-1}} \bx_0 + \sqrt{1-\alpha_{t-1}-\sigma_t^2}\cdot \frac{\bx_t-\sqrt{\alpha_t} \bx_0}{\sqrt{1-\alpha_t}}, \sigma_t^2 \mathbf{I}). \nonumber 
\end{align}
where $\{\alpha_t\}_{t=0}^T$ is a predefined increasing sequence, and $\sigma_t$ controls the stochasticity of the diffusion trajectory.
Since the distribution of $\bx_t$ given $\bx_0$ is modeled as $q_{\sigma}(\bx_t \mid \bx_0) = \mathcal{N}(\sqrt{\alpha_t} \bx_0, (1 - \alpha_t)\mathbf{I})$, the ground truth $\bx_0$ can be approximated using Tweedie's formula~\cite{stein1981estimation} as:
\begin{equation}
    \hat{\bx}_0 (\bx_t, t) \coloneqq \frac{\bx_t-\sqrt{1-\alpha_t} \cdot \boldsymbol{\epsilon}_\theta (\bx_t, t)}{\sqrt{\alpha_t}},
\end{equation}
where $\boldsymbol{\epsilon}_\theta(\cdot, \cdot)$ is a noise prediction network, typically implemented using a U-Net~\cite{ronneberger2015u} or Transformer architecture~\cite{peebles2023scalable}.
Generally, we set $\sigma_t = 0$ which makes the deterministic DDIM reverse process as
\begin{align}
    \bx_{t-1} & = \sqrt{\frac{\alpha_{t-1}}{\alpha_t}}\bx_t+(1-\alpha_t)\gamma_t\nabla_{\bx_t}\log{p(\bx_t)} \nonumber \\
    & \approx \sqrt{\frac{\alpha_{t-1}}{\alpha_t}}\bx_t-\sqrt{1-\alpha_t}\gamma_t\epsilon_\theta(\bx_t, t)
    \label{eq:org_update}
\end{align}
where $\gamma_t := \sqrt{\alpha_{t-1}/\alpha_t} - \sqrt{(1-\alpha_{t-1})/(1-\alpha_t)}$.

\subsubsection{Stochastic differential equation}
As shown in Song \textit{et al.}~\cite{song2023score}, the noise perturbation in DDPM\cite{ho2020denoising} and DDIM~\cite{song2020denoising} can be modeled as a stochastic process governed by a discretized forward SDE~\cite{kloeden1992stochastic}:
\begin{align}
    \mathrm{d} \bx = f(\bx, t)\mathrm{d}t + g(t)\mathrm{d}\mathbf{w},
\end{align}
which has a corresponding reverse SDE that denotes the reverse process of the diffusion model as follows:
\begin{align}
    \mathrm{d}\bx = [f(\bx, t)-g(t)^2\nabla_\bx \log{p_t(\bx)}]\mathrm{d}t + g(t)\mathrm{d}\bar{\mathbf{w}}.
\end{align}

\subsubsection{Notations}

Before introducing our method, we define the notations used throughout the paper.
Let $\{\mathbf{X}_t\}_{t=0}^T$ denote the original objective we aim to generate.
We define $\{f_i\}_{i=1}^N$ as a set of mapping functions that project $\mathbf{X}_t$ into $N$ different patches $\{\mathbf{y}_t^i\}_{i=1}^N$, where $\mathbf{y}_t^i \coloneqq f_i (\mathbf{X}_t)$.
Since each $\by_t^i$ has a resolution compatible with the diffusion model, we apply the diffusion process to each patches, resulting in diffusion trajectories $\{\by_t^i\}_{t=0}^T$ for each $i$.

For example, in the case of wide image generation (Sec.~\ref{subsubsec:wide_image}), $\mathbf{X}_t$ corresponds to the wide image itself, and $f_i$ is a cropping function that extracts patch $\by_t^i$ from $\mathbf{X}_t$.
In contrast, for 3D mesh texturing (Sec.~\ref{subsubsec:3d_mesh_texturing}), $\mathbf{X}_t$ represents the texture map of an input 3D mesh, and $f_i$ transforms $\mathbf{X}_t$ into a rendered image of the mesh from a specific viewpoint.
Additionally, we define $\tilde{\mathbf{X}}^i \coloneqq \cup_{j=1}^{i-1} \{\by_t^j\}_{t=1}^T$ and $\tilde{\mathbf{X}}_t^i \coloneqq \cup_{j=1}^{i-1} \by_t^j$.
We elaborate on how the union is defined for each task in the following sections.

\begin{figure}[t]\centering
\includegraphics[width=0.95\linewidth]{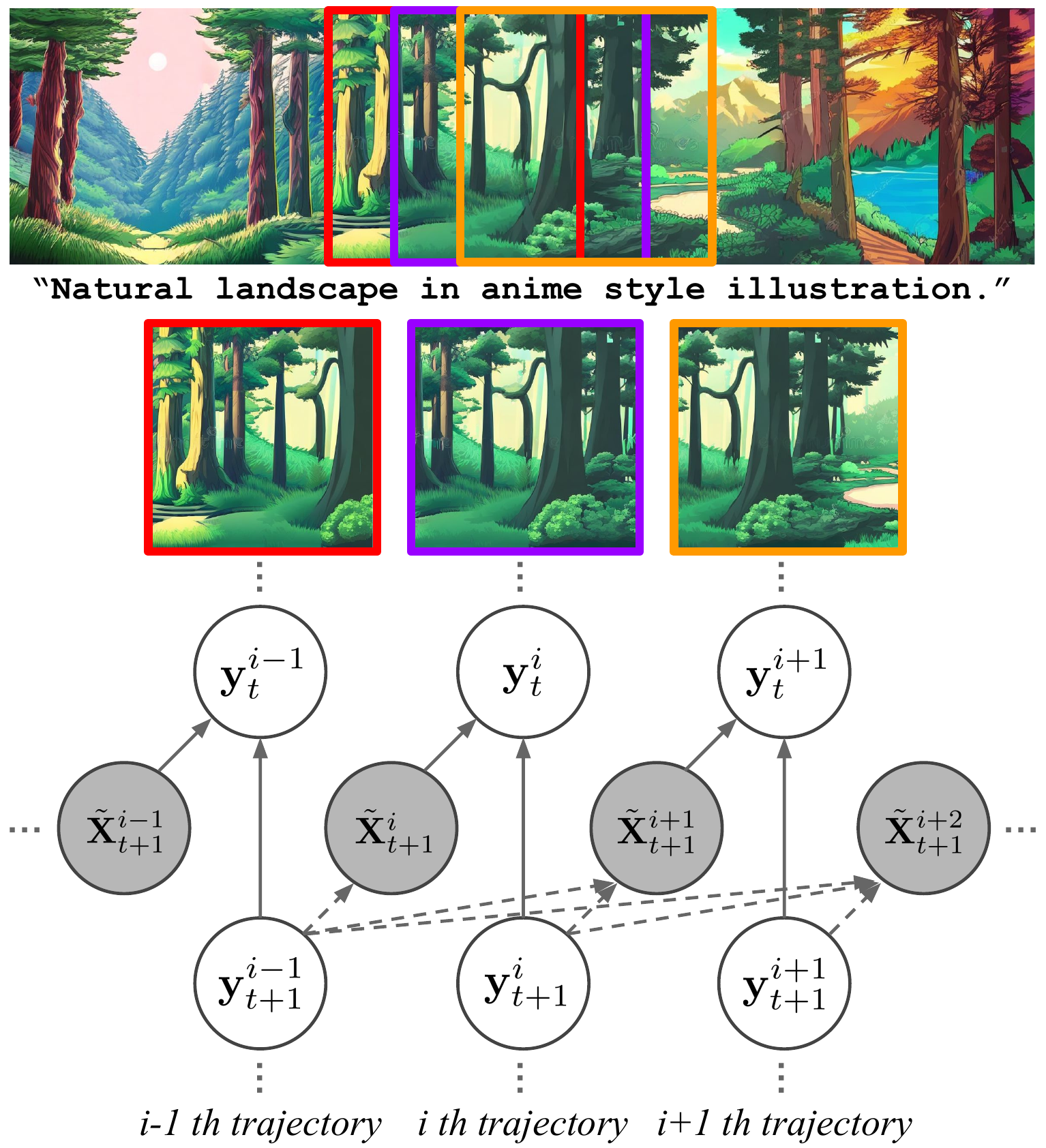}
\vspace{-2mm}
\captionof{figure}{\textbf{
Graphical diagram of our method.} 
We sequentially generate trajectories conditioned on previous generations. 
}
\label{fig:method}
\vspace{-3mm}
\end{figure}

\subsection{Proposed framework: SyncSDE}
\label{subsec:method_overview}
We propose a synchronous generation process that sequentially generates the trajectories of $\{\by_t^i\}_{t=0}^T$.  
First, we generate the trajectory $\{\by_t^1\}_{t=0}^T$.  
Then, $\{\by_t^2\}_{t=0}^T$ is generated, conditioned on the previously generated trajectory $\{\by_t^1\}_{t=0}^T$.  
This process continues iteratively, where each trajectory $\{\by_t^i\}_{t=0}^T$ is conditioned on the previously generated trajectories $\{\by_t^1\}_{t=0}^T, \{\by_t^2\}_{t=0}^T, \ldots, \{\by_t^{i-1}\}_{t=0}^T$.  
A graphical illustration of the proposed generation process is shown in Fig.~\ref{fig:method}.
We model the relationship between trajectories only at the same timestep, \ie, for all $t_1 \neq t_2$ and $i \neq j$, we assume $\by_{t_1}^i \indep \by_{t_2}^j$.  
Inspired by the conditional score estimation proposed in~\cite{lee2024conditional}, we derive the score function for the conditional generation process as follows:
\begin{align}
    & \nabla_{\by_t^i} \log{p(\by_t^i \mid \tilde{\mathbf{X}}^i}) =\nabla_{\by_t^i} \log{p(\by_t^i \mid \tilde{\mathbf{X}}_t^i)} \nonumber \\
    &= \nabla_{\by_t^i} \log{p(\by_t^i)} + \nabla_{\by_t^i} \log{p(\tilde{\mathbf{X}}_t^i \mid \by_t^i)} .
    \label{eq:conditional_score_fn}
\end{align}
The first term of Eq.~\ref{eq:conditional_score_fn} corresponds to the original score function estimated by the pretrained diffusion model.  
The second term captures the relationship between generations, which we explicitly model.  
By substituting Eq.~\ref{eq:conditional_score_fn} into Eq.~\ref{eq:org_update}, the reverse SDE sampling update for the $i^{\text{th}}$ view at timestep $t$ is given by:
\begin{align}
    \by_{t-1}^i =& \sqrt{\frac{\alpha_{t-1}}{\alpha_{t}}}\by_{t}^i+(1-\alpha_{t})\gamma_{t}\nabla_{\by_{t}^i}\log{p(\by_{t}^i \mid \tilde{\mathbf{X}}^i}) \nonumber \\
    \approx& \sqrt{\frac{\alpha_{t-1}}{\alpha_{t}}}\by_{t}^i-\sqrt{1-\alpha_{t}} \gamma_{t} \epsilon_\theta(\by_{t}^i, t) \nonumber \\
    &+(1-\alpha_{t})\gamma_{t}\nabla_{\by_{t}^i}\log{p(\tilde{\mathbf{X}}_{t}^i \mid \by_{t}^i)}.
    \label{eq:update_eq}
\end{align}
Note that we focus on modeling the heuristic to estimate $\nabla_{\by_t^i} \log{p(\tilde{\mathbf{X}}_t^i \mid \by_t^i)}$.  
Our approach significantly reduces the search space for empirical testing when identifying optimal diffusion synchronization strategies.  
Following the setup in~\cite{lee2024conditional}, we model this conditional probability term for each task using a single tunable hyperparameter, $\lambda$.

\subsection{Applications of SyncSDE}
\label{subsec:image_generation_syncsde}

We explore a range of collaborative generation tasks to demonstrate the applicability of SyncSDE, including mask-based text-to-image generation, text-driven real image editing, wide image generation, ambiguous image generation, 3D mesh texturing, and long-horizon motion generation.

\subsubsection{Mask-based Text-to-Image generation}
\label{subsubsec:mask_generation}
For mask-based text-to-image generation, we use three prompts:  
$p^{\mathrm{bg}}$ for the background, $p^{\mathrm{fg}}$ for the masked region, and $p^{\mathrm{img}}$ for the overall image that semantically integrates $p^{\mathrm{bg}}$ and $p^{\mathrm{fg}}$.  
We generate the final image using three variables:  
$\by_t^1$ for the background conditioned on $p^{\mathrm{bg}}$,  
$\by_t^2$ for the masked region using $p^{\mathrm{fg}}$,  
and $\by_t^3$ to refine the entire image based on $p^{\mathrm{img}}$.
Our goal is to generate a high-quality image $\mathbf{X}$, defined as $\by_0^3$.  
We first generate $\by_t^1$, then synthesize $\by_t^2$—representing the foreground region—using the conditional score function defined in Eq.~\ref{eq:conditional_score_fn}.
The conditional distribution is defined as:
\begin{equation}
    p(\tilde{\mathbf{X}}_t^2 \mid \by_t^2) \sim \mathcal{N}(\by_t^2, \lambda(1-\alpha_t)  \mathbf{M}^{-1}),
    \label{eq:mask_t2i_prob1}
\end{equation}
where $\tilde{\mathbf{X}}_t^2$ is equal to ${\by}_t^1$, $\mathbf{M}$ is a diagonal precision matrix indicating the background region of the image, and $\lambda$ is a tunable hyperparameter.  
The diagonal elements of $\mathbf{M}$ are constructed as:
\begin{equation}
    \mathrm{Diag}(\mathbf{M}) = \mathrm{Reshape}(\mathbf{B}),
    \label{eq:mask_reshape}
\end{equation}
where $\mathbf{B} \in \mathbb{R}^{H \times W}$ is a binary mask such that $\mathbf{B}[h, w] \in \{0, 1\}$ indicates background (1) and foreground (0) regions.  
The $\mathrm{Reshape}(\cdot)$ operation flattens the matrix from $\mathbb{R}^{H \times W}$ to a vector in $\mathbb{R}^{HW}$.  
The intuition behind Eq.~\ref{eq:mask_t2i_prob1} is that foreground pixels require higher variance to allow object formation, while background pixels should maintain lower variance to ensure consistency.  
Accordingly, we assign smaller variance to the background and larger variance to the foreground.
We finally generate $\by_t^3$ using the following conditional probability:
\begin{align}
    p(\tilde{\mathbf{X}}_t^3 \mid \by_t^3) & \sim  \mathcal{N}(\by_t^3, \lambda(1-\alpha_t) \mathbf{M}^{-1}) \nonumber \\ & \quad \cdot  \mathcal{N}(\by_t^3, \lambda(1-\alpha_t) (\mathbf{1}-\mathbf{M})^{-1}),
    \label{eq:mask_t2i_prob2}
\end{align}
where $\mathbf{1}$ is a diagonal precision matrix with all diagonal elements set to 1. Note that $\tilde{\mathbf{X}}_t^3$ is computed as:
\begin{equation}
   \tilde{\mathbf{X}}_t^3 = \mathbf{M} \odot \by_t^1 + (\mathbf{1} - \mathbf{M}) \odot \by_t^2,
    \label{eq:mask_t2i_aux}
\end{equation}
where $\odot$ is the Hadamard product.
We choose $\mathbf{X} = \by_0^3$.

\subsubsection{Text-driven real image editing}
\label{subsubsec:i2i_translation}
The text-driven real image editing task requires precise modifications, as it aims to manipulate only the foreground region while preserving the background.  
Given a source image $\mathbf{x}^\mathrm{src}$, a source prompt $p^\mathrm{src}$, and a target prompt $p^\mathrm{tgt}$, we first invert the source image using the forward SDE~\cite{kloeden1992stochastic} to obtain the latent sequence $\{ \mathbf{x}^\mathrm{src}_t \}_{t=0}^{T}$.  
Following CSG~\cite{lee2024conditional}, we generate a soft mask $\tilde{\mathbf{B}}$ that identifies the background region of the source image using attention maps~\cite{vaswani2017attention} from the pretrained diffusion model.  
Details of the soft mask generation process are provided in the Appendix.  
We then obtain the binary mask $\mathbf{B}$ as follows:
\begin{equation}
    \mathbf{B}[h, w] = \chi \left( \tilde{\mathbf{B}}[h, w] \geq \tau \right),
    \label{eq:i2i_mask_generation}
\end{equation}
where $\chi$ outputs 1 if the given condition is true, and 0 otherwise, and $\tau \in [0, 1]$ is a threshold for attention values.  
Following the logic of mask-based T2I generation, we generate the target image $\mathbf{x}^{\mathrm{tgt}}$.  
We first apply the binary mask obtained from Eq.~\ref{eq:i2i_mask_generation} to Eq.~\ref{eq:mask_reshape} to construct $\mathbf{M}$.  
Next, we replace each $\by_t^1$ with $\bx_t^{\mathrm{src}}$, and set $\by_T^3 = \bx_T^{\mathrm{src}}$.  
Finally, using $p^{\mathrm{tgt}}$, we sample $\{ \by_t^2 \}_{t=0}^T$ and $\{ \by_t^3 \}_{t=0}^{T-1}$, and obtain the edited image as $\mathbf{x}^{\mathrm{tgt}} = \by_0^3$.

\subsubsection{Wide image generation}
\label{subsubsec:wide_image}
To generate a wide image, we define the operation $f_i$ as a cropping function that extracts the image patch $\by_t^i$ from the wide image $\mathbf{X}_t$.  
The patches $\{ \by_0^i \}_{i=1}^{N}$ are defined to be \textit{partially overlapped}.
We then design the conditional probability term as follows:
\begin{equation}
    p(\tilde{\mathbf{X}}_t^i \mid \by_t^i) \sim \mathcal{N}(\by_t^i, \lambda(1 - \alpha_t) \mathbf{M}_i^{-1}),
    \label{eq:wide_image_prob}
\end{equation}
where
\begin{equation}
    \tilde{\mathbf{X}}_t^i = (\mathbf{1} - \mathbf{M}_i) \odot f_i(f_{i-1}^{-1}(\by_t^{i-1})),
    \label{eq:wide_image_aux}
\end{equation}
and $\mathbf{M}_i$ is a binary mask that indicates the non-overlapping pixels between the $i^{\mathrm{th}}$ and $(i{-}1)^{\mathrm{th}}$ patches.
After generating all $N$ patches, we apply an overlapping operation $\varphi$ to combine them into the initial reconstruction $\mathbf{X}_0$:
\begin{equation}
    \mathbf{X}_0 = \varphi \left( \{ f_i^{-1}(\by_0^i) \}_{i=1}^N \right).
\end{equation}
The operation $\varphi$ ensures that patches with larger $i$ values are placed on top in overlapping regions.  
Finally, we decode $\mathbf{X}_0$ using the VAE decoder~\cite{kingma2013auto} of LDM~\cite{rombach2022high} to obtain the final wide image $\mathbf{X}$.

\subsubsection{Ambiguous image generation}
\label{subsubsec:ambiguous_image}

An ambiguous image is designed to support multiple interpretations through visual transformations $f_i$. These transformations include operations such as identity mapping, (counter) clockwise rotation, skewing, and flipping. The specific types of $f_i$ used in our experiments are described in Sec.~\ref{sec:exp_ambiguous}.  
We define the conditional probability as:
\begin{equation}
    p(\tilde{\mathbf{X}}_t^i \mid \by_t^i) \sim \mathcal{N}(\by_t^i, \lambda(1 - \alpha_t)\mathbf{1}),
    \label{eq:ambiguous_image_prob}
\end{equation}
where $\tilde{\mathbf{X}}_t^i$ is defined as
\begin{equation}
    \tilde{\mathbf{X}}_t^i = f_i(f_{i-1}^{-1}(\by_t^{i-1})),
    \label{eq:ambiguous_image_aux}
\end{equation}
and $\lambda$ is a tunable hyperparameter.

\subsubsection{3D mesh texturing}
\label{subsubsec:3d_mesh_texturing}
For 3D mesh texturing, we define the variable $\by^i$ as the projected image of a 3D mesh observed from the $i^{\mathrm{th}}$ viewpoint.  
We then design the conditional probability term as:
\begin{equation}
    p(\tilde{\mathbf{X}}_t^i \mid \by_t^i) \sim \mathcal{N}(\by_t^i, \lambda(1 - \alpha_t) \mathbf{M}_i^{-1}),
    \label{eq:3d_mesh_prob}
\end{equation}
where $\lambda$ is a tunable hyperparameter, and $\tilde{\mathbf{X}}_t^i$ is defined as:
\begin{equation}
    \tilde{\mathbf{X}}_t^i = f_i(f_{i-1}^{-1}(\{ \by_t^j \}_{j=1}^{i-1}))[i].
    \label{eq:3d_mesh_aux}
\end{equation}
Here, $\mathbf{M}_i$ is a binary mask indicating the background region of the $i^{\mathrm{th}}$ view, generated during the rendering process.  
The function $f_{i-1}^{-1}$ is an inverse-projection function that composes a texture map from images captured at the first $i{-}1$ viewpoints, while $f_i$ is a projection function that renders the texture map into $i$ viewpoint-specific images.  
We take the $i^{\mathrm{th}}$ image from the output of $f_i$ to obtain $\tilde{\mathbf{X}}_t^i$.

\subsubsection{Long-horizon motion generation}
\label{subsubsec:motion_generation}

For long-horizon motion generation, we generate short-duration motion segments with MDM~\cite{tevet2023human} with overlapping timestamps to smoothly form an extended, coherent motion sequence. We define the operation $f_i$ as a query function for extracting motion segment $\mathbf{y}_t^i$ from the total motion sequence $\mathbf{X}_t$. These segments, $\{{\mathbf{y}_0^i}\}_{i=1}^{N}$, are constructed to have \textit{partial temporal overlaps}. To achieve coherent transitions, we define the conditional probability as:
\begin{equation}
p(\tilde{\mathbf{X}}_t^i \mid \mathbf{y}_t^i) \sim \mathcal{N}(\mathbf{y}_t^i, \lambda(1-\alpha_t)\mathbf{M}_i^{-1}),
\label{eq:long_motion_prob}
\end{equation}
where
\begin{equation}
\tilde{\mathbf{X}}_t^i = (\mathbf{1} - \mathbf{M}_i) \odot f_i(f_{i-1}^{-1}(\mathbf{y}_t^{i-1})),
\label{eq:long_motion_aux}
\end{equation}
and $\mathbf{M}_i$ is a binary mask indicating non-overlapping timestamps between the $i^{\mathrm{th}}$ and $(i-1)^{\mathrm{th}}$ motion segments.
After generating $N$ motion segments, we combine them using an overlapping operation $\varphi$, ensuring smooth continuity at overlapping timestamps, as follows:
\begin{equation}
\mathbf{X}_0 = \varphi(\{{f_i^{-1}(\mathbf{y}_0^i)}\}_{i=1}^{N}).
\end{equation}
The operation $\varphi$ prioritizes later segments (larger $i$) in regions where timestamps overlap, ensuring temporal consistency in the complete long-horizon motion sequence $\mathbf{X}$.

\begin{table}[t]
    \caption{
    \textbf{Quantitative results of mask-based T2I generation.}
    We generate images using the pretrained Stable Diffusion~\cite{rombach2022high}. KID score is scaled by $10^3$.
    \vspace{-2mm}
    }
    \small
    \centering
    \resizebox{0.95\columnwidth}{!}{
        \begin{tabular}{lccc}
            \toprule
            Method & KID~\cite{binkowski2018demystifying} $\downarrow$& FID~\cite{heusel2017gans} $\downarrow$ &  CLIP-S~\cite{radford2021learning} $\uparrow$ \\
            \midrule
            MultiDiffusion~\cite{omer2023multidiffusion}& 
            47.694 & 84.225 & 0.330 \\
            SyncTweedies~\cite{jaihoon2024synctweedies}& 
            117.360 & 149.470 & 0.307 \\
            SyncSDE ($1/\lambda=5$)  & \cellcolor{tabsecond}{43.774} & \cellcolor{tabsecond}{82.878} & \cellcolor{tabfirst}{0.332} \\
            SyncSDE (best)  & \cellcolor{tabfirst}{34.859} & \cellcolor{tabfirst}{72.118} & \cellcolor{tabsecond}{0.331} \\
            \bottomrule
        \end{tabular}
    }
    \label{tab:mask_generation}
\end{table}

\begin{figure}[t]\centering
\includegraphics[width=0.9\linewidth]{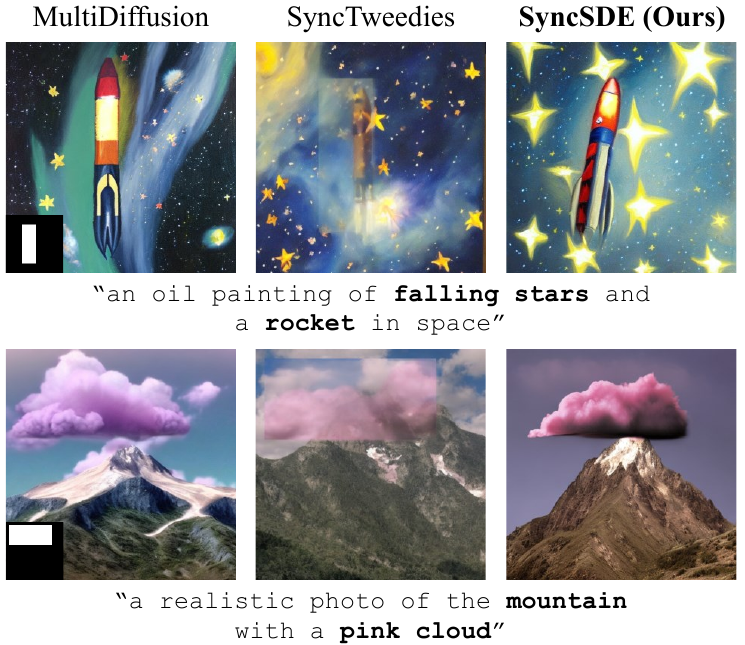}
\captionof{figure}{\textbf{
Qualitative results of mask-based T2I generation.} 
We find that SyncSDE effectively models the correlation between background and foreground giving comparable result with task specific model, MultiDiffusion~\cite{omer2023multidiffusion}, while SyncTweedies~\cite{jaihoon2024synctweedies} fails and tends to generate blurry image in the masked region.
}
\label{fig:mask_generation}
\vspace{-3mm}
\end{figure}

\section{Experiments}
\label{sec:exp}

In this section, we qualitatively and quantitatively evaluate the performance of SyncSDE.  
We compare it against SyncTweedies~\cite{jaihoon2024synctweedies} across tasks presented in Sec.~\ref{subsubsec:mask_generation}$\sim$\ref{subsubsec:3d_mesh_texturing}, and against task-specific methods~\cite{omer2023multidiffusion, daniel2024visual, liu2024text, su2022dual, meng2021sdedit, cao2023masactrl, dong2023prompt} for text-driven image editing in Sec.~\ref{subsubsec:i2i_translation}.  
We then present results for long-horizon motion generation (Sec.~\ref{subsubsec:motion_generation}), highlighting the scalability of our approach.  
Finally, we analyze the impact of the hyperparameter $\lambda$.

\subsection{Implementation details}
\label{sec:exp_implementation}

We implement our method based on the official codebases of CSG\footnote{https://github.com/Hleephilip/CSG}~\cite{lee2024conditional} and SyncTweedies\footnote{https://github.com/KAIST-Visual-AI-Group/SyncTweedies}~\cite{jaihoon2024synctweedies}.  
All experiments are conducted using the DDIM~\cite{song2020denoising} sampler as the numerical solver for the reverse SDE.  
For fair comparison, we use the same number of DDIM sampling steps across all methods and tasks.  
We also apply classifier-free guidance~\cite{ho2022classifier} with a consistent guidance scale across all baselines.
For ease of implementation, we use $1/\lambda$ in place of $\lambda$, and employ a scheduler that linearly decreases $1/\lambda$ as the timestep $t$ decreases.  
Task-specific details are provided in the following subsections and in the Appendix.

\begin{table}[t]
    \caption{
    \textbf{Quantitative results of text-driven real image editing.} 
    We use real images from LAION-5B dataset~\cite{schuhmann2022laion} and the pretrained Stable Diffusion~\cite{rombach2022high}.
    SyncSDE shows better performance compared to task-specific methods~\cite{su2022dual, meng2021sdedit, dong2023prompt, cao2023masactrl}.
    \vspace{-2mm}
    }
    \centering
    \resizebox{1.0\columnwidth}{!}{
        \begin{tabular}{lcccc}
            \toprule
            Method & CLIP-S~\cite{radford2021learning} $\uparrow$ & LPIPS~\cite{zhang2018perceptual} $\downarrow$ & BG-LPIPS~\cite{zhang2018perceptual} $\downarrow$  \\
            \midrule
            DDIB~\cite{su2022dual}& 
            0.294 & 0.379  & 0.350  \\
            SDEdit~\cite{meng2021sdedit}  & 0.298 & 0.407 & 0.369  \\
            PTI~\cite{dong2023prompt} & \cellcolor{tabfirst}{0.322} & 0.409 & 0.379  \\
            MasaCtrl~\cite{cao2023masactrl} & 0.285 & {0.290} & 0.341 \\
            SyncSDE ($1/\lambda=5$) & 0.311 & \cellcolor{tabsecond}{0.281} & \cellcolor{tabsecond}{0.266} \\
            SyncSDE (best) & \cellcolor{tabsecond}{0.313} & \cellcolor{tabfirst}{0.254} & \cellcolor{tabfirst}{0.222} \\
            \bottomrule
        \end{tabular}
    }
    \vspace{-2mm}
    \label{tab:i2i_translation}
\end{table}

\begin{figure}[t]\centering
\includegraphics[width=0.9\linewidth]{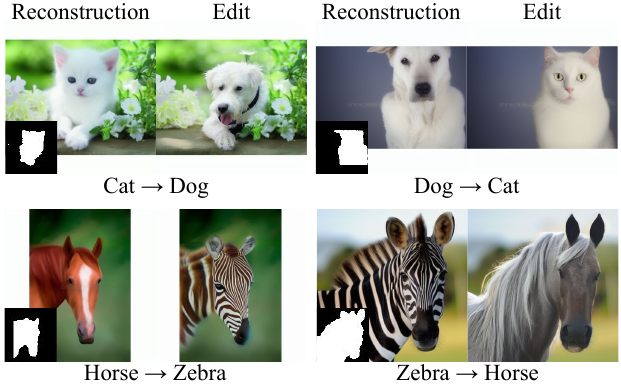}
\captionof{figure}{\textbf{
Qualitative results of text-driven real image editing.} We visualize the results of various real image editing tasks using the real images sampled from LAION-5B dataset~\cite{schuhmann2022laion}. SyncSDE shows a good performance on text-driven image manipulation.
}
\label{fig:real_image_editing}
\end{figure}

\subsection{Results}
\label{subsec:exp_results}

We present quantitative and/or qualitative results of the proposed method and baseline algorithms~\cite{jaihoon2024synctweedies, omer2023multidiffusion, daniel2024visual, liu2024text, su2022dual, meng2021sdedit, dong2023prompt, cao2023masactrl} across the collaborative generation tasks discussed in Section~\ref{subsec:image_generation_syncsde}.  
For SyncSDE, we report results using two different values of $\lambda$: one with a fixed setting of $1/\lambda = 5$, which performs well across tasks, and another which is tuned for each task to achieve the best results.
In all tables, we highlight the best (\raisebox{0.6ex}{\colorbox{tabfirst}{\hspace{1.5mm}\vspace{0.7mm}}}) and second-best (\raisebox{0.6ex}{\colorbox{tabsecond}{\hspace{1.5mm}\vspace{0.7mm}}}) results in each column using colored cells.  
Additional results and implementation details are provided in the Appendix.

\subsubsection{Mask-based Text-to-Image generation}

We compare the proposed method with SyncTweedies~\cite{jaihoon2024synctweedies} and MultiDiffusion~\cite{omer2023multidiffusion} using the pretrained Stable Diffusion~\cite{rombach2022high}.
Since the official implementation of SyncTweedies does not include mask-basd T2I generation, we modified the code to produce the results. 
We generate two diffusion trajectories ($\mathbf{w}_0, \mathbf{w}_1$), representing the background and masked region, respectively.
Final output image $\mathbf{z}$ is computed by using $\mathbf{w}_0$ as the background while averaging the masked region with $\mathbf{w}_1$ as follows:
\begin{equation}
    \mathbf{z} = \mathbf{M} \odot \mathbf{w}_0 + (\mathbf{1}
 - \mathbf{M}) \odot (\mathbf{w}_0 + \mathbf{w}_1)
\end{equation}
where $\mathbf{M}$ is the background mask explained in Section~\ref{subsubsec:mask_generation}, $\odot$ is Hadamard product operator. We follow the notation of SyncTweedies for variable $\mathbf{z}$ and $\mathbf{w}$ which are denoted as canonical and instance variable respectively.

For quantitative evaluations, we use  KID~\cite{binkowski2018demystifying} and FID~\cite{heusel2017gans} to quantify the fidelity of the generated images, with CLIP-S~\cite{radford2021learning} to measure the similarity between the generated images and the given text prompts.
As shown in Table~\ref{tab:mask_generation}, SyncSDE significantly outperforms SyncTweedies and MultiDiffusion across all three metrics.
Also, SyncSDE generates plausible images that seamlessly composites foreground with the background, as illustrated in Figure~\ref{fig:mask_generation}. 
In contrast, SyncTweedies struggles with localization and synchronization, failing to blend the object into the overall image. 
While MultiDiffusion relies on an additional bootstrapping strategy specifically for object localization, SyncSDE achieves superior performance without any extra components, highlighting the effectiveness of our method.

\begin{table}[t]
    \caption{
    \textbf{Quantitative results of wide image generation.}
    We generate wide images using the pretrained Stable Diffusion~\cite{rombach2022high}.
    Note that KID score is scaled by $10^3$.
    \vspace{-2mm}
    }
    \small
    \centering
    \resizebox{0.95\columnwidth}{!}{
        \begin{tabular}{lccc}
            \toprule
            Method & KID~\cite{binkowski2018demystifying} $\downarrow$& FID~\cite{heusel2017gans} $\downarrow$ &  CLIP-S~\cite{radford2021learning} $\uparrow$ \\
            \midrule
            SyncTweedies~\cite{jaihoon2024synctweedies}& 
            51.024 & 78.333  & \cellcolor{tabfirst}{0.328} \\
            SyncSDE ($1/\lambda=5$)  & \cellcolor{tabsecond}{17.311} & \cellcolor{tabsecond}{44.969} & \cellcolor{tabsecond}{0.324} \\
            SyncSDE (best)  & \cellcolor{tabfirst}{16.872} & \cellcolor{tabfirst}{44.707} & \cellcolor{tabsecond}{0.324} \\
            \bottomrule
        \end{tabular}
    }
    \label{tab:wide_image_generation}
\end{table}

\begin{figure}[t]\centering
\includegraphics[width=\linewidth]{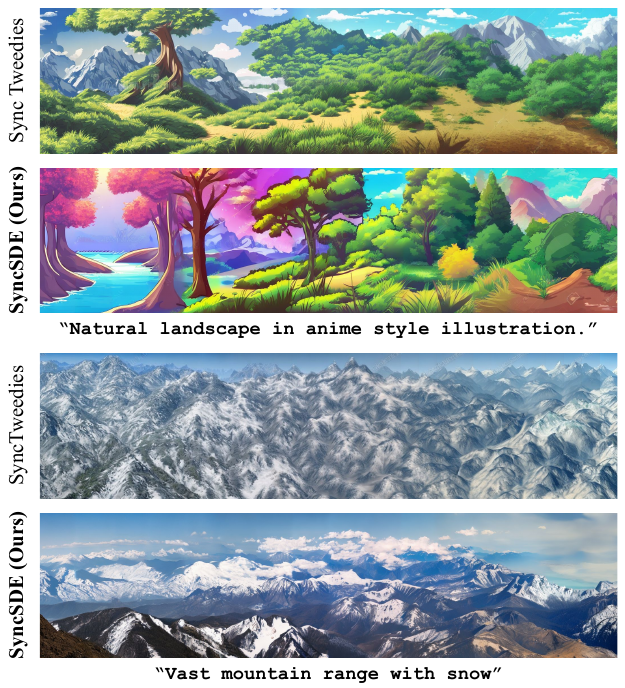}
\captionof{figure}{\textbf{
Qualitative results of wide image generation.} We visualize $2048 \times 512$ sized wide image generated by our method and SyncTweedies~\cite{jaihoon2024synctweedies}.
SyncSDE generates better quality of wide images in a continuous manner compared to SyncTweedies.
}
\label{fig:wide_image}
\vspace{-2mm}
\end{figure}

\subsubsection{Text-driven real image editing}
We compare our method with prior works~\cite{su2022dual, meng2021sdedit, dong2023prompt, cao2023masactrl} with the pre-training stable diffusion~\cite{rombach2022high}.
Since the official implementation of PTI~\cite{dong2023prompt} is unavailable, we reproduce it using the same initial latent $\bx_T^{\mathrm{src}}$.
For comparisons, we sample 1,000 real images from the LAION-5B dataset~\cite{schuhmann2022laion}. 
We generate the source prompt using the pretrained image captioning model BLIP~\cite{li2022blip}.
Then, the target prompt corresponding to the edited image is generated by swapping the words of the source prompt. 
We follow the quantitative evaluation protocol of Pix2Pix-Zero~\cite{pix2pixzero}.
We evaluate each methods with CLIP-S score~\cite{radford2021learning} to measure the similarity between the edited image and the target prompt.
In addition, we measure LPIPS~\cite{zhang2018perceptual} to quantify the perceptual similarity between the source and the edited image.
To further evaluate background preservation, we calculate LPIPS using the background region (BG-LPIPS), which is segmented using the pretrained image segmentation model Detic~\cite{zhou2022detecting}.
As shown in Table~\ref{tab:i2i_translation} and Figure~\ref{fig:real_image_editing}, SyncSDE shows superior performance in text-driven real image editing.
Note that `Reconstruction' denotes the reconstructed source image $\by^1_0$, while `Edit' means the edited image $\by^3_0$.

\begin{table}[t]
    \caption{
    \textbf{Quantitative results of ambiguous image generation.}
    We generate ambiguous images using the pretrained Deepfloyd~\cite{DeepFloydIF}.
    Note that KID score is scaled by $10^3$.
    \vspace{-2mm}
    }
    \small
    \centering
    \resizebox{0.95\columnwidth}{!}{
        \begin{tabular}{lccc}
            \toprule
            Method & KID~\cite{binkowski2018demystifying} $\downarrow$& FID~\cite{heusel2017gans} $\downarrow$ &  CLIP-S~\cite{radford2021learning} $\uparrow$ \\
            \midrule
            Visual Anagrams~\cite{daniel2024visual} & 195.286 & 215.082 & \cellcolor{tabfirst}{0.290} \\
            SyncTweedies~\cite{jaihoon2024synctweedies}& 
            215.119 & 226.922 & 0.262 \\
            SyncSDE ($1/\lambda=5$)  & \cellcolor{tabfirst}{173.590} & \cellcolor{tabsecond}{212.196} & \cellcolor{tabsecond}{0.273} \\
            SyncSDE (best)  & \cellcolor{tabsecond}{174.902} & \cellcolor{tabfirst}{208.788} & 0.272 \\
            \bottomrule
        \end{tabular}
    }
    \label{tab:ambiguous_image_generation}
\end{table}

\begin{figure}[t]\centering
\includegraphics[width=0.95\columnwidth]{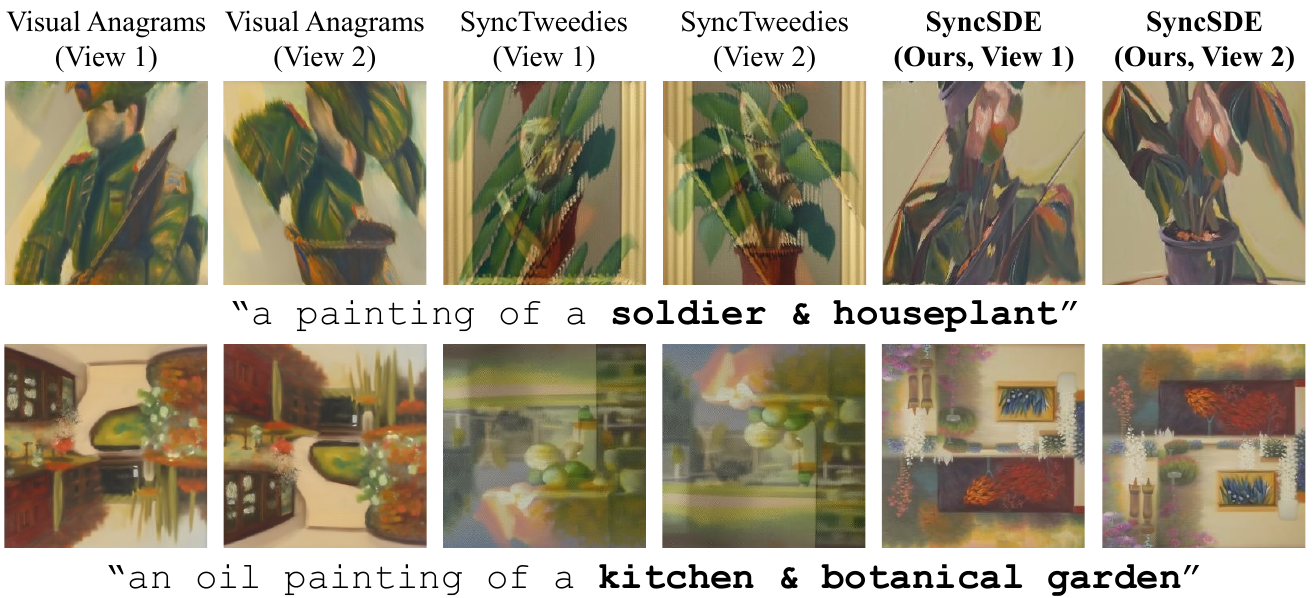}
\captionof{figure}{\textbf{
Qualitative results of ambiguous image generation.} 
We visualize the ambiguous images generated by SyncSDE, SyncTweedies~\cite{jaihoon2024synctweedies}, and Visual Anagrams~\cite{daniel2024visual}. The first row applies identity and skew transformations, while the second row applies identity and flip transformations.
SyncSDE generates realistic images that blends in both prompts, while SyncTweedies fails to integrate two prompts.
}
\label{fig:ambiguous_image_1}
\end{figure}

\subsubsection{Wide image generation}
We generate $2048 \times 512$ resolution wide image using the pretrained Stable Diffusion~\cite{rombach2022high} as backbone.
Quantitative results of wide image generation is presented in Table~\ref{tab:wide_image_generation}.
Our method outperforms SyncTweedies~\cite{jaihoon2024synctweedies} in terms of KID~\cite{binkowski2018demystifying}, FID~\cite{heusel2017gans}, and CLIP-S score~\cite{radford2021learning}.
Additionally, Figure~\ref{fig:wide_image} shows that our method produces more plausible and high-fidelity results compared to SyncTweedies.
Notably, in the wide image generated with the prompt ``Vast mountain range with snow'', SyncTweedies fails to synthesize realistic views, highlighting the limitations of patch averaging.

\begin{table}[t]
    \caption{
    \textbf{Quantitative results of 3D mesh texturing.}
    We generate textures using the pretrained ControlNet~\cite{zhang2023adding}.
    Note that KID score is scaled by $10^3$.
    \vspace{-2mm}
    }
    \small
    \centering
    \resizebox{0.95\columnwidth}{!}{
        \begin{tabular}{lccc}
            \toprule
            Method & KID~\cite{binkowski2018demystifying} $\downarrow$& FID~\cite{heusel2017gans} $\downarrow$ &  CLIP-S~\cite{radford2021learning} $\uparrow$ \\
            \midrule
            SyncMVD~\cite{liu2024text} & 188.328 & 183.663 & \cellcolor{tabfirst}{0.317} \\
            SyncTweedies~\cite{jaihoon2024synctweedies}& \cellcolor{tabsecond}{186.648} & \cellcolor{tabsecond}{183.387} & {0.311}\\
            
            SyncSDE ($1/\lambda = 5$)  & {187.373} & {184.873} & \cellcolor{tabsecond}{0.312} \\
            
            SyncSDE (best)  & \cellcolor{tabfirst}{169.631} & \cellcolor{tabfirst}{174.756} & \cellcolor{tabsecond}{0.312} \\
            \bottomrule
        \end{tabular}
    }
    \label{tab:mesh_texturing}
\end{table}

\begin{figure}[t]\centering
\includegraphics[width=0.9\linewidth]{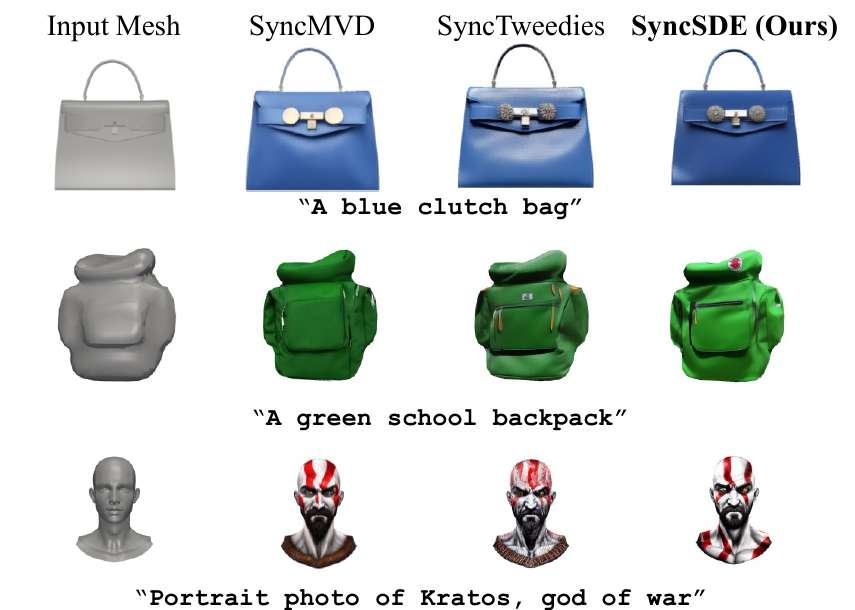}
\captionof{figure}{\textbf{
Qualitative result of 3D mesh texturing.} We qualitatively compare the performance of the proposed method on 3D mesh texturing with SyncMVD~\cite{liu2024text} and SyncTweedies~\cite{jaihoon2024synctweedies}.
SyncSDE gives high-quality textured images corresponding to the given text prompt.
}
\label{fig:3D_texture}
\end{figure}

\subsubsection{Ambiguous image generation}
\label{sec:exp_ambiguous}

We compare our method with SyncTweedies~\cite{jaihoon2024synctweedies} and Visual Anagrams~\cite{daniel2024visual}. 
To generate ambiguous images, we use the pretrained DeepFloyd-IF~\cite{DeepFloydIF}.
We generate a single image using two prompts, modeling two semantics within the image by choosing $f_1$ and $f_2$.
We fix $f_1$ as identity mapping, and choose $f_2$ from 4 types of transformation explained in Section~\ref{subsubsec:ambiguous_image}; (1) $\pm 90^\circ$ rotation, (2) $180^\circ$ rotation, (3) vertical flip, and (4) skew transformation.
Note that skew transformation shifts columns of image pixels with offset.

Table~\ref{tab:ambiguous_image_generation} shows that SyncSDE outperforms all baselines across KID~\cite{binkowski2018demystifying} and FID~\cite{heusel2017gans} scores, and has comparable CLIP-S score~\cite{radford2021learning}.
Figure~\ref{fig:ambiguous_image_1} further illustrates that our method generates significantly better results compared to prior works.
Especially, SyncTweedies tend to generate blurry images that appear as simple averages of the two different images from each views, rather than plausibly blended images.
In some cases, the objects given in two different text prompts become unidentifiable.
We claim that this issues is due to the lack of theoretical foundation behind patch averaging. 
In contrast, SyncSDE generates more coherent images, where views are seamlessly blended and properly correlated.
Additionally, we emphasize that a significant benefit of our method is its ability to produce high-quality results compared to Visual Anagrams, despite Visual Anagrams being specifically designed only for ambiguous image generation. 

\subsubsection{3D mesh texturing}
\label{sec:exp_mesh_texturing}
We compare the proposed method with SyncTweedies~\cite{jaihoon2024synctweedies} and SyncMVD~\cite{liu2024text}.
Note that we utilize the pretrained depth-conditioned ControlNet~\cite{zhang2023adding} as backbone architecture.
We synchronize 10 different diffusion processes, to generate a texture for a given 3D mesh.
Eight of the processes correspond to views with azimuth values evenly spaced by $45^\circ$ within the range $[0^\circ, 360^\circ)$, and two auxiliary trajectories encoding views with azimuth values $0^\circ$ and $180^\circ$, both at an elevation of $30^\circ$. 
Each diffusion process is modeled with ControlNet~\cite{zhang2023adding} conditioned on depth information extracted from the input mesh. 
The results are then compared using the projected images of the generated texture map.

Table~\ref{tab:mesh_texturing} shows that SyncSDE outperforms prior works in terms of KID~\cite{binkowski2018demystifying}, FID~\cite{heusel2017gans}, and CLIP-S score~\cite{rombach2022high}.
As shown in Figure~\ref{fig:3D_texture}, our method qualitatively surpasses the performance of the baselines, while SyncTweedies tend to blur the details of the texture.~\footnote{\textbf{Note:} The results reported in the arXiv version differ from those in the official camera-ready version of CVPR. 
We identified several implementation errors and have updated the results after fixing them.
\textbf{Despite these changes, the overall tendency of the results remains consistent, hence we maintain the thesis.}
Detailed explanations of the errors and the corresponding revisions are provided in Appendix~\ref{sec:supp_mesh} and on \href{https://hjl1013.github.io/SyncSDE/}{project page}.}

\begin{figure}[t]\centering
\includegraphics[width=\linewidth]{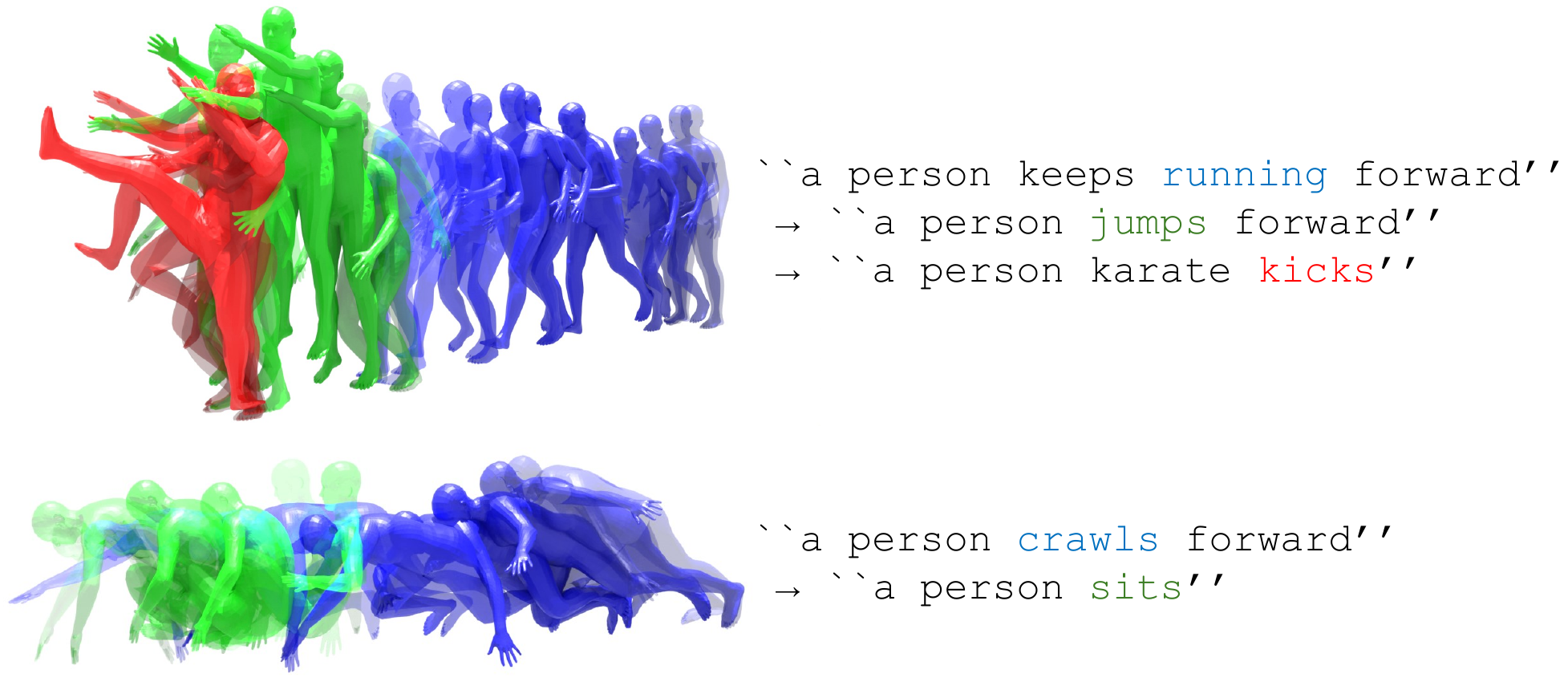}
\captionof{figure}{
\textbf{Qualitative results of long-horizon motion generation.}  
Our method generates coherent long-horizon motion sequences by synchronizing multiple output trajectories from a motion diffusion model~\cite{tevet2023human}, where each trajectory produces a short motion segment.
}
\label{fig:motionfig}
\vspace{-2mm}
\end{figure}

\subsubsection{Long-horizon motion generation}
We demonstrate the broad applicability of SyncSDE through long-horizon motion generation, as shown in Fig.~\ref{fig:motionfig}.  
Specifically, we use the motion diffusion model~\cite{tevet2023human}, where each trajectory generates a short human motion sequence of 120 frames.  
To compose a continuous motion, we set $1/\lambda = 3$ and apply a 0.25 overlap ratio across timesteps (\ie, 30 frames overlap).  
SyncSDE successfully synchronizes the motion segments, producing a coherent long-horizon sequence with smooth transitions between segments.

\subsection{Effects of $\lambda$}

We analyze the effect of the hyperparameter $\lambda$ by generating ambiguous images using different values of $\lambda$.
Theoretically, $\lambda$ controls the degree of collaboration between multiple diffusion trajectories.
In the probabilistic formulation of $p(\tilde{\mathbf{X}}_t^i \mid \by_t^i)$ in Eq.~\ref{eq:ambiguous_image_prob}, a smaller value of $\lambda$ reduces the variance of the $N^\mathrm{th}$ trajectory relative to the $1^\mathrm{st} \sim (N-1)^\mathrm{th}$ trajectories, thereby preserving semantic features encoded in $1^\mathrm{st} \sim (N-1)^\mathrm{th}$ trajectories.
In contrast, larger $\lambda$ values increase variance, allowing the $N^\mathrm{th}$ trajectory to deviate more and depend less on earlier trajectories.

\begin{figure}[t]
\centering
\includegraphics[width=\linewidth]{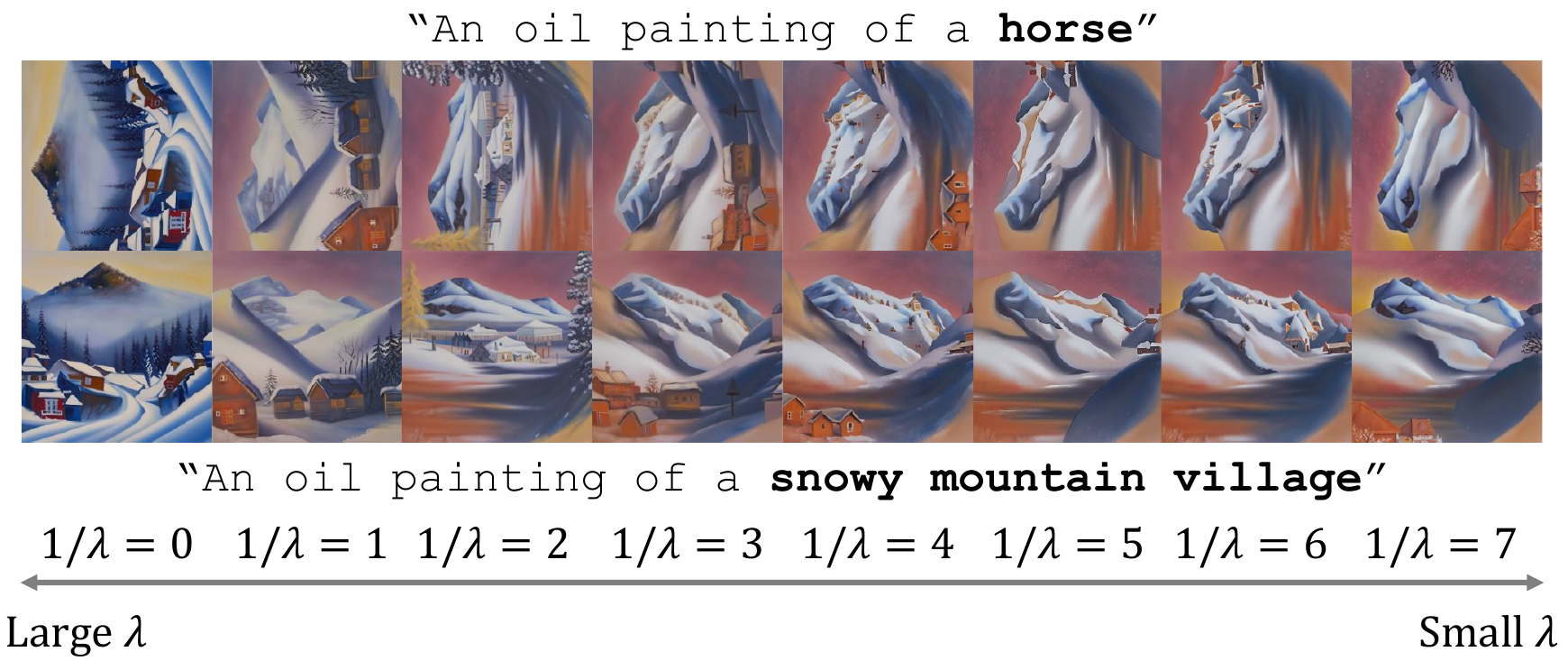}
\caption{\textbf{Effects of $\lambda$.} We show the balance of collaboration between two prompts.
When $\lambda$ is large, the effect of $2^{\mathrm{nd}}$ prompt dominates, while the effect of $1^{\mathrm{st}}$ prompt becomes significant for smaller $\lambda$ value.
}
\label{fig:qual_lambda_ambiguous}
\end{figure}

This theoretical analysis is visually supported by Figure ~\ref{fig:qual_lambda_ambiguous}. 
When $1/\lambda = 0$ (\ie, $\lambda=\infty$), the $2^\mathrm{nd}$ trajectory becomes independent of the $1^\mathrm{st}$, generating an image that only aligns with the $2^{\mathrm{nd}}$ prompt. 
Mathematically, this eliminates the conditional probability term in Eq.~\ref{eq:ambiguous_image_prob}, resulting in a failure of integrating both prompts, which is consistent with our theoretical explanation.
As $1/\lambda$ increases (\ie, $\lambda$ decreases), trajectory correlation strengthens, generating a well-blended image incorporating both prompts.
If $1/\lambda$ further increases, the balance between two trajectories collapses, causing the $1^{\mathrm{st}}$ prompt to dominate.
This effect is clearly visualized with $1/\lambda=6$ and $1/\lambda=7$, where the features of the horse dominate the features of the mountain village.

\section{Conclusion}
\label{sec:conclusion}

We propose a probabilistic framework for diffusion synchronization, providing a theoretical analysis of why it works.
By designing conditional probabilities between diffusion trajectories, we establish synchronization across multiple trajectories.
Based on the proposed method, we focus on efficient heuristic modeling by identifying which probability term to model, significantly reducing the empirical testing to find optimal solutions. 
We evaluate our method on various collaborative generation tasks, comparing its performance with prior works. 
Experimental results demonstrate that our method is widely applicable and consistently outperforms baseline algorithms. 
We hope this work inspires future research on more robust and principled models of inter-trajectory correlations to further advance diffusion synchronization.

\section*{Acknowledgment}
This paper was supported by Korea Institute for Advancement of Technology (KIAT) grant funded by the Korea Government (Ministry of Education) (P0025681-G02P22450002201-10054408, ``Semiconductor''-Specialized University)

{
    \small
    \bibliographystyle{ieeenat_fullname}
    \bibliography{main}
}

\clearpage


\setcounter{equation}{23}
\setcounter{figure}{9}
\setcounter{table}{5}
\appendix

\section{Task-specific experimental details}
\label{sec:appendix_method}

In this section, we provide the experimental details of each diffusion synchronization task.

\subsection{Mask-based Text-to-Image generation}

We use the pretrained Stable Diffusion v2 checkpoint~\cite{rombach2022high} for image generation, resulting in $512 \times 512$ resolution image.
Using 10 prompts, we generate 250 images per prompt with a fixed background mask for each.
KID~\cite{binkowski2018demystifying} and FID~\cite{heusel2017gans} we use 2,000 images per prompt using the same pretrained model.
We use 50 steps for DDIM~\cite{song2020denoising} sampling.

\subsection{Text-driven real image editing}

Firstly, we explain the details of soft mask generation.
Note that we follow CSG~\cite{lee2024conditional} to generate the soft mask $\tilde{\mathbf{B}}$ which indicates the background region of the source image.
Following paragraph summarizes the procedure introduced in~\cite{lee2024conditional}.

We extract the self-attention and cross-attention map of the source image using the pretrained Stable Diffusion~\cite{rombach2022high}, each denoted as $\mathbf{M}_{\mathrm{self}} \in \mathbb{R}^{H \times W \times H \times W}$ and $\mathbf{M}_{\mathrm{cross}} \in \mathbb{R}^{N \times H \times W}$, where $N$ denotes the number of word tokens defined in the pretrained Stable Diffusion model.
Then we generate the background mask $\tilde{\mathbf{B}}$ as follows:
\begin{equation}
    \tilde{\mathbf{B}} = \mathbf{1} - \mathbf{M}_{\mathrm{fg}},
\end{equation}
where each element of $\mathbf{M}_{\mathrm{fg}} \in \mathbb{R}^{H \times W}$ is defined as
\begin{equation}
    \mathbf{M}_\mathrm{fg}[h, w] = \text{tr}(\mathbf{M}_{\mathrm{self}}[h, w] \mathbf{M}_{\mathrm{cross}}^{\top}[u]).
\end{equation}
Note that $u$ denotes the index of the word token corresponds to the object that we want to manipulate.

We use the pretrained Stable Diffusion v1-4 model for experiments, generate images in $512 \times 512$ resolution.
Also, we use four image editing tasks for evaluation: cat $\rightarrow$ dog, dog $\rightarrow$ cat, horse $\rightarrow$ zebra, and zebra $\rightarrow$ horse.
For each task, we sample 250 real images from the LAION-5B dataset~\cite{schuhmann2022laion}.
To find the most relevant images for the source word (\eg `cat' in cat-to-dog task) within the dataset, we leverage CLIP retrieval~\cite{beaumont-2022-clip-retrieval}.
The source prompt is generated using the pretrained BLIP model~\cite{li2022blip}, while the target prompt is constructed by replacing the source word with the target word.
For instance, in the `horse $\rightarrow$ zebra' task, we swap the word `horse' in the source prompt with `zebra' to generate the target prompt.
We use DDIM~\cite{song2020denoising} sampling with 50 steps.

\subsection{Wide image generation}

We use the pretrained Stable Diffusion v2 checkpoint~\cite{rombach2022high} for wide image generation.
With four different text prompts, we generate 250 images per prompt at a resolution of $2048 \times 512$.
To measure KID~\cite{binkowski2018demystifying} and FID~\cite{heusel2017gans}, we randomly crop the generated wide images to a resolution of $ 512 \times 512$.
For CLIP-S score~\cite{radford2021learning}, we use a center crop operation.
We generate 2,000 images per prompt to construct the reference image set using the same pretrained model.
We use 50 steps for DDIM~\cite{song2020denoising} sampling.

\subsection{Ambiguous image generation}
We use the pretrained DeepFloyd v1.0 checkpoint~\cite{DeepFloydIF} for experiments, synthesizing images at $256 \times 256$ resolution.
The DeepFloyd-IF model employs a two-stage sampling process for image generation. 
Note that we apply the proposed synchronization startegy only to the $1^{\mathrm{st}}$ stage, while the $2^{\mathrm{nd}}$ stage's sampling is performed without synchronization.
We use 5 prompt pairs, where each pair consists of two prompts describing the semantics to be modeled in resulting ambiguous image.
For each prompt pair, we set $f_1$ as identity mapping and choose $f_2$ from one of 4 visual transforms: $\pm 90^\circ$ rotation, $180^\circ$ rotation, vertical flip, and skew transformation. 
We then generate 250 images per prompt pair.
In case of reference images for measuring KID~\cite{binkowski2018demystifying} and FID~\cite{heusel2017gans}, we generate 2,000 images per prompt in each prompt pair with the same pretrained model.
Total 30 timesteps are used for DDIM~\cite{song2020denoising} sampling.

\begin{figure}[t]\centering
\includegraphics[width=\linewidth]{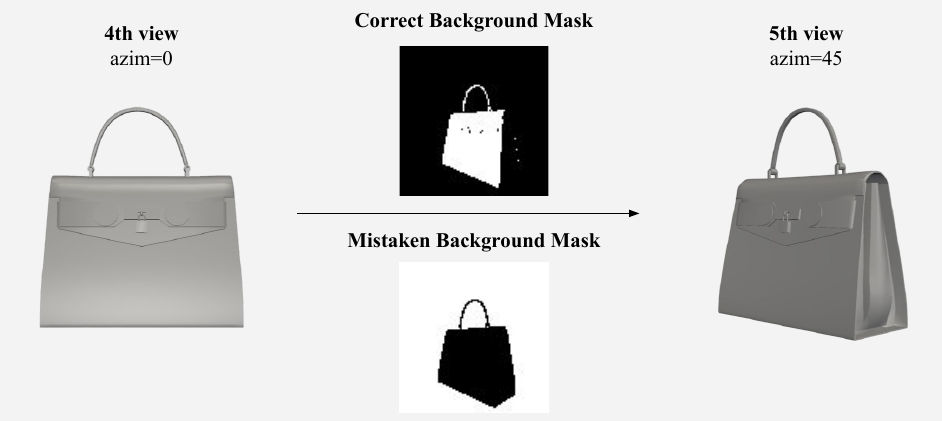}
\captionof{figure}{\textbf{
Revision of the background mask in SyncSDE.} 
We visualize the corrected background mask, where background part is shown is denoted as white pixels.
}
\label{fig:mesh_mask_revision}
\end{figure}

\begin{figure*}[t]\centering
\includegraphics[width=0.95\linewidth]{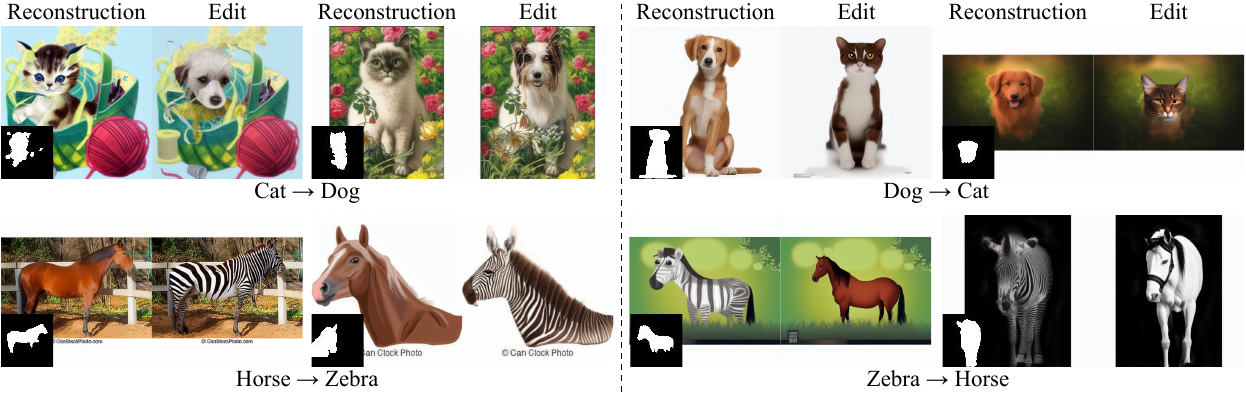}
\captionof{figure}{\textbf{
Additional qualitative results of text-driven real image editing.} 
We edit the real images sampled from the LAION-5B dataset~\cite{schuhmann2022laion} by leveraging SyncSDE combined with the pretrained Stable Diffusion~\cite{rombach2022high}. 
We also visualize the foreground region defined by the generated mask.
}
\label{fig:i2i_supp}
\end{figure*}

\subsection{3D mesh texturing}
\label{sec:supp_mesh}
We use the pretrained depth-conditioned ControlNet v1-1~\cite{zhang2023adding} for mesh texturing.
Using 6 meshes and a single prompt for each mesh, we generate 100 textures per mesh.
Each generated texture is projected onto a fixed single view, resulting in a $768 \times 768$ resolution RGB image.
To generate reference images, we use the same pretrained model and sample 2,000 images per prompt using the equivalent mesh map as depth condition.
In addition, following SyncMVD~\cite{liu2024text} and SyncTweedies~\cite{jaihoon2024synctweedies}, we use the self-attention modification technique proposed in~\cite{liu2024text} along with Voronoi Diagram-guided filling~\cite{aurenhammer1991voronoi}.
We use 30 steps for DDIM~\cite{song2020denoising} sampling.
Like SyncTweedies, we do not use diffusion synchronization during the last 20\% of the sampling steps.

\noindent \paragraph{Revisions.}
As mentioned in Section~\ref{sec:exp_mesh_texturing}, we identified several implementation errors that affected the reported results of SyncSDE and SyncMVD~\cite{liu2024text}.
We now present the revised results after correcting these issues.

For SyncSDE, the error came from the background mask used in Eq.~\ref{eq:3d_mesh_prob}. 
Our intention was to define the background mask as the region already filled by previous views during the autoregressive texture generation process. 
However, we mistakenly used a static background mask instead, as illustrated in Figure~\ref{fig:mesh_mask_revision}.
For SyncMVD, we found inconsistencies in the configuration of the UV texture map and the visualization settings compared to SyncTweedies~\cite{jaihoon2024synctweedies} and SyncSDE, leading to an unfair comparison.
Specifically, we revised the latent texture size from 512 to 1536 and the RGB view resolution from 1536 to 768 to match the settings used in SyncTweedies and SyncSDE. 
We also replaced orthographic cameras with perspective cameras and updated the mesh normalization from $L_\infty$ normalization to $L_2$ normalization to ensure consistency across all baselines.

Although these modifications led to changes in the quantitative results (Table~\ref{tab:mesh_texturing}), the overall trend remains unchanged, thus preserving the validity of our claims regarding the effectiveness of our method.

\begin{table}[h]
    \caption{
    \textbf{Comparison of computational overhead between SyncSDE and SyncTweedies~\cite{jaihoon2024synctweedies}.
    }
    }
    \small
    \centering
    \resizebox{\columnwidth}{!}{
        \begin{tabular}{lcc}
            \toprule
            Method & Time (s/image) & GPU memory (GB) \\
            \midrule
            SyncTweedies~\cite{jaihoon2024synctweedies}  & 7.721 & 2.44 \\
            SyncSDE (Ours) & 5.664 & 2.78 \\
            \bottomrule
        \end{tabular}
    }
    \label{tab:exp_comp_overhead}
    \vspace{-5mm}
\end{table}

\section{Computational overhead}

We analyze the computational overhead in terms of both time and GPU memory required to generate a single image.
The measured computational overhead of the proposed method and SyncTweedies~\cite{jaihoon2024synctweedies} is reported in Table~\ref{tab:exp_comp_overhead}.
We use a single NVIDIA RTX 3090 GPU for measurement. 
Notably, SyncSDE exhibits a comparable computational overhead to SyncTweedies.

\section{Additional qualitative results}
\label{sec:appendix_qual}

\begin{figure*}[t]\centering
\includegraphics[width=0.95\linewidth]{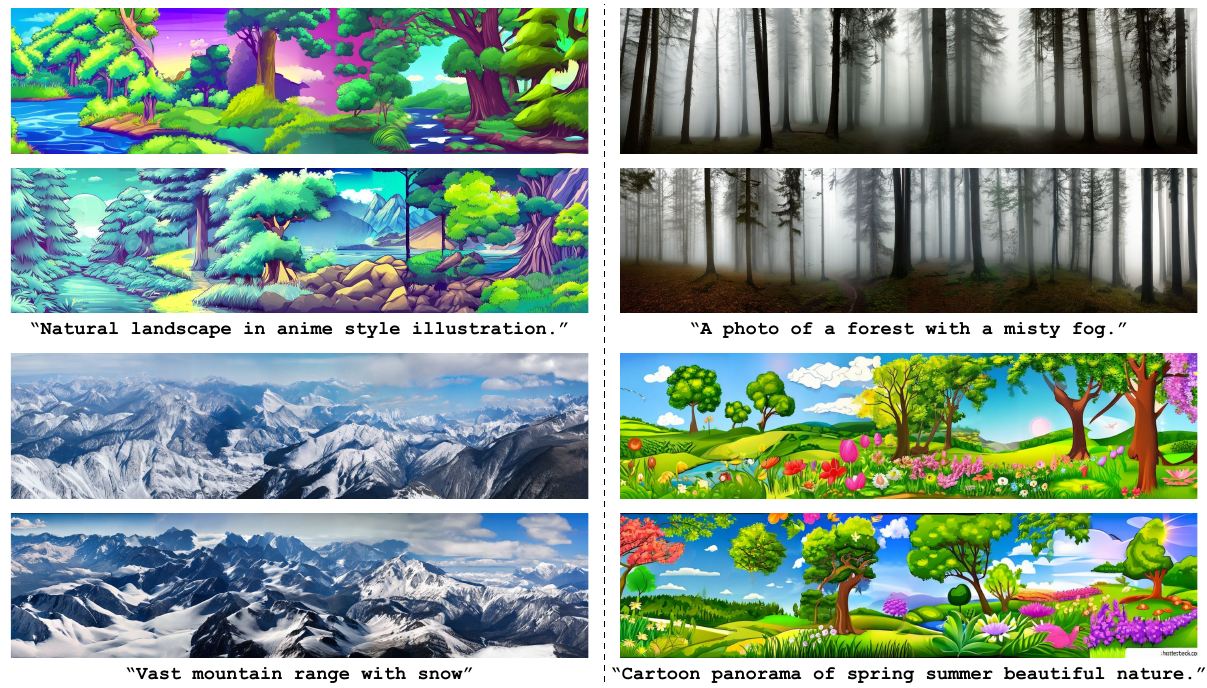}
\captionof{figure}{\textbf{
Additional qualitative results of wide image generation.} 
We visualize wide images generated by SyncSDE using the pretrained Stable Diffusion~\cite{rombach2022high} for image generation.
}
\label{fig:wide_image_supp}
\end{figure*}

\begin{figure*}[t]\centering
\includegraphics[width=0.95\linewidth]{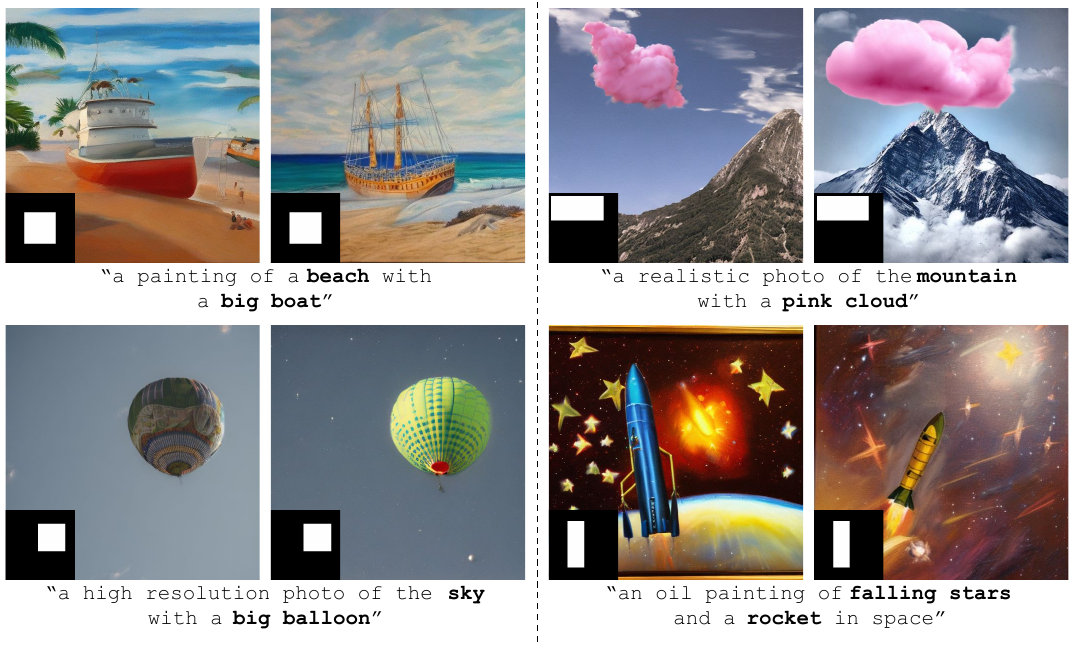}
\captionof{figure}{\textbf{
Additional qualitative results of mask-based T2I generation.} 
SyncSDE shows strong performance on mask-based T2I generation task. We use the pretrained Stable Diffusion~\cite{rombach2022high} for image generation.
}
\label{fig:mask_generation_supp}
\end{figure*}

\begin{figure*}[t]\centering
\includegraphics[width=0.95\linewidth]{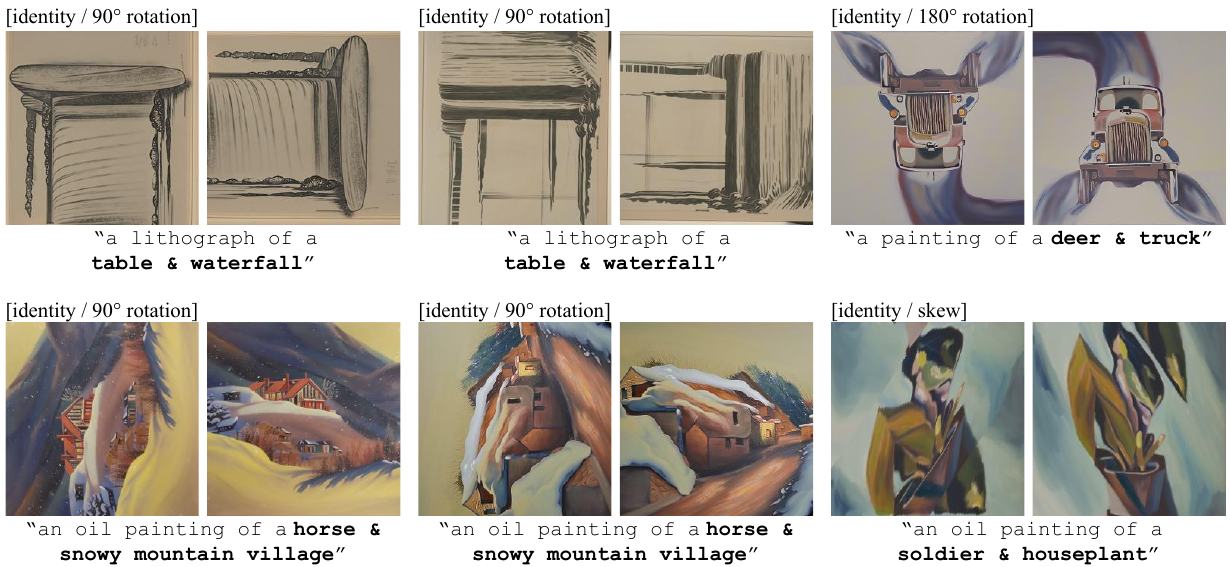}
\captionof{figure}{\textbf{
Additional qualitative results of ambiguous image generation.} 
Using the pretrained Deepfloyd-IF model~\cite{DeepFloydIF}, we generate ambiguous image with various prompt pairs and visual transformations. 
SyncSDE generates high-quality ambiguous images.
}
\label{fig:ambiguous_image_supp}
\end{figure*}

\begin{figure*}[t]\centering
\includegraphics[width=0.95\linewidth]{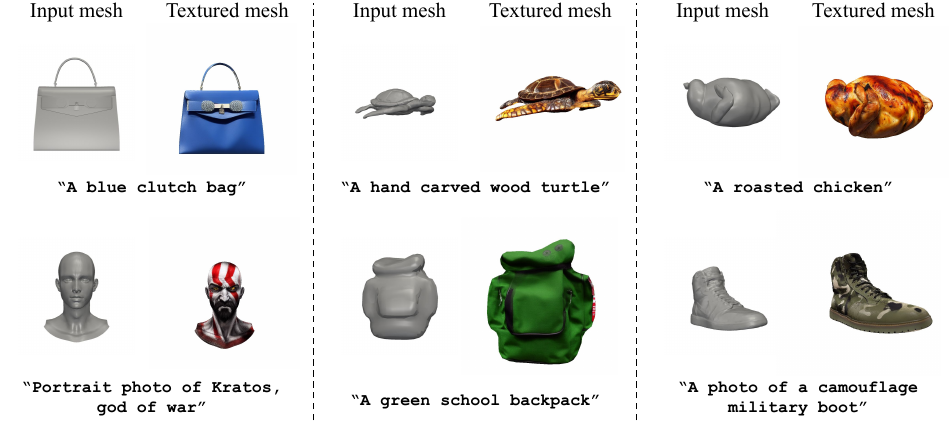}
\captionof{figure}{\textbf{
Additional qualitative results of 3D mesh texturing.} 
We use the pretrained depth-conditioned ControlNet~\cite{zhang2023adding} for mesh texturing.
Given an input mesh and the text prompt, SyncSDE generates remarkable texture images.
}
\label{fig:mesh_texuring_supp}
\end{figure*}

We visualize additional qualitative results of SyncSDE in Figure~\ref{fig:i2i_supp},~\ref{fig:wide_image_supp},~\ref{fig:mask_generation_supp},~\ref{fig:ambiguous_image_supp}, and~\ref{fig:mesh_texuring_supp}.
As shown in the figures, SyncSDE shows outstanding performance in multiple image generation tasks, including mask-based T2I generation, text-driven real image editing, wide image generation, ambiguous image generation, and 3D mesh texturing.
The experimental results demonstrate that the proposed method successfully models the correlation between multiple diffusion trajectories, thus smoothly blending the generated patches.

\section{Limitations and social impacts}
\label{sec:appendix_limitations}

Since our method uses a pretrained text-to-image diffusion model~\cite{rombach2022high, DeepFloydIF, zhang2023adding}, the proposed method may result in suboptimal outcomes depending on the pretrained backbone model.
For instance, due to the limitations of the pretrained diffusion model, it may struggle to synthesize images with complex structures or multiple fine details.
Furthermore, the proposed method may generate harmful images due to the shortcomings of the pretrained diffusion model.

\clearpage

\end{document}